\documentclass{article}

\usepackage[sglblindworkshop, final]{latinx_2025}

\usepackage[utf8]{inputenc} 
\usepackage[T1]{fontenc}    
\usepackage{hyperref}       
\usepackage{url}            
\usepackage{booktabs}       
\usepackage{amsfonts}       
\usepackage{nicefrac}       
\usepackage{microtype}      
\usepackage{xcolor}         
\usepackage{tikz}
\usetikzlibrary{arrows.meta, positioning}
\usepackage{enumitem}

\hypersetup{
  hidelinks,
  unicode,
  breaklinks=true,
  pdfauthor={Joao P. A. Dantas},
  pdftitle={Offline Language-Based Support System for Tactical Decision-Making in Air Combat Operations}
}

\title{Offline Language-Based Support System for Tactical Decision-Making in Air Combat Operations}

\author{
  Joao~P.~A.~Dantas\thanks{\url{www.joaopadantas.com}} \\
  Decision Support Systems Subdivision\\
  Institute for Advanced Studies\\
  São José dos Campos, São Paulo, Brazil, 12228-001 \\
  \texttt{jpdantas@ita.br} \\
}

\begin{document}

\maketitle

\begin{abstract}
An offline language-based support tool is presented to assist pilot training and tactical decision-making in air combat operations. The system, named MATE (Mission Assistant for Training and Evaluation), applies retrieval-augmented generation (RAG) to answer natural language questions using content extracted from doctrinal and tactical manuals. Built entirely with open-source components, it performs text preprocessing, embedding generation, semantic search, and language generation locally, ensuring full offline functionality. The tool was evaluated by three Brazilian Air Force fighter pilots, who rated the answers produced with three open-weight language models (Llama 3.1 8B, Qwen 2.5 14B, and Mistral NeMo 12B) to five tactical questions derived from a publicly available air combat manual. Average ratings ranged from 4.00 to 4.40 on a five-point scale, with Mistral NeMo 12B obtaining the highest score, and the pilots rated the response time of the tool as fully adequate compared to manual search. The relevance of the retrieved content was identified as the main point for improvement. Potential applications are discussed in onboard mission systems, flight simulators, and training environments.
\end{abstract}

\section{Introduction}

Air combat missions often require fast and reliable access to tactical information that is usually found in lengthy and complex manuals. During planning or even in flight, pilots and mission planners may need to consult this content quickly. However, doing so manually is often slow and impractical, especially in environments where internet access is not available or allowed. The ability to interact with technical documents in a natural and efficient way is increasingly valuable in defense contexts~\citep{doll2024natural}.

Recent progress in natural language processing (NLP), especially with large language models (LLMs), has made it possible to build systems that answer user questions in plain language using information from documents~\citep{brown2020language, openai2023gpt4}. However, most of these tools depend on cloud services, which limits their use in sensitive or disconnected environments such as military aircraft or secure simulators.

To address this gap, we present an offline language-based support tool that uses retrieval-augmented generation (RAG)~\citep{lewis2020retrieval} to provide helpful answers based on doctrinal manuals. The system combines document search with a local language model, allowing users to ask questions in natural language and receive relevant, context-aware answers. It is built entirely with open-source tools and works fully offline, making it suitable for use in onboard systems, simulators, and training environments. The system is modular and can be adapted to different types of documents.

The main contribution of this work is the design and implementation of a tactical tool that runs completely offline and helps users access sensitive documents that should not be processed online. We show how it works using real examples from air combat manuals and explain how it can support decision-making and training in air combat situations. We also report an expert validation in which Brazilian Air Force fighter pilots rated the quality of the answers produced by the tool with three different open-weight language models.

\section{Related Work}
\label{sec:related_work}

NLP has been increasingly applied in military contexts to enhance decision-making, intelligence analysis, and operational efficiency~\citep{teze2024future}. Early initiatives, such as DARPA's TIPSTER program, aimed to improve information retrieval and extraction for defense applications, laying the groundwork for subsequent advancements in military NLP systems~\citep{darpa_tipster}.

Recent developments have seen the integration of NLP in various military operations. For instance, DARPA's Transformative Apps (TransApps) program developed and rapidly fielded secure mobile applications for deployed military units, showing how software tools can be delivered directly to field operations~\citep{transapps}. Similarly, the MiTAP system, developed by the Mitre Corporation, utilizes NLP to gather, translate, and present information for monitoring global events, aiding in biosecurity and defense intelligence~\citep{mitap}.

In the field of intelligence analysis, NLP has been employed to process vast amounts of unstructured data. Primer, an AI company, has developed NLP tools for the U.S. Special Operations Command to analyze and identify disinformation campaigns, showcasing the role of NLP in modern information warfare~\citep{primer_disinfo}. Additionally, the COA-GPT model employs large language models to generate courses of action in military operations, highlighting the potential of NLP in strategic planning~\citep{coa_gpt}.

Applications of AI and language interfaces in air combat have also been explored. For example, the Air Combat Evolution (ACE) program by DARPA aims to develop collaboration between AI and human pilots, allowing pilots to act as ``battle managers'' for unmanned aircraft~\citep{darpa_ace}. Moreover, speech recognition systems have been implemented in advanced fighter aircraft, such as the F-35 Lightning II, to allow pilots to manage subsystems such as communications and navigation by voice, thereby reducing pilot workload~\citep{schutte2007f35speech}.

\section{Methodology}
\label{sec:methodology}

The proposed system, MATE (Mission Assistant for Training and Evaluation), is designed to provide offline, natural language access to tactical knowledge by combining document retrieval and generative language modeling. It follows a modular architecture with five key elements: document indexing, semantic retrieval, language model interaction, user interface, and offline deployment.

\textbf{Document Indexing:} The pipeline starts by preprocessing and indexing doctrinal and tactical manuals stored locally. Supported formats (.txt, .pdf, .docx) are converted into plain text and split into overlapping segments of 1,200 characters, with an overlap of 250 characters between consecutive segments. The multilingual \texttt{paraphrase-multilingual-MiniLM-L12-v2} model from SentenceTransformers~\citep{reimers2019sentence, reimers2020multilingual} encodes each segment into a 384-dimensional embedding. The embeddings are L2-normalized and stored in a FAISS inner-product index~\citep{johnson2019billion}, so that the similarity search corresponds to cosine similarity. A metadata file links each segment to its source document.

\textbf{Semantic Retrieval:} When a user submits a natural language question, the system computes its embedding using the same model and performs a nearest-neighbor search over the FAISS index. The top-$k$ most similar segments ($k=5$) are selected and concatenated, in order of similarity, into a textual context limited to 4,000 characters, which is passed as input to the language model together with the question. All retrieval logic is implemented in Python using the FAISS and SentenceTransformers libraries.

\textbf{Language Model Interaction:} The context and user query are forwarded to a locally deployed LLM via the Ollama API~\citep{ollama}, which handles model loading, memory management, and inference entirely offline. Any open-weight model available in Ollama can be used; in this work, the tool was evaluated with Llama 3.1 8B~\citep{grattafiori2024llama3}, Qwen 2.5 14B~\citep{qwen2024qwen25}, and Mistral NeMo 12B~\citep{mistral_nemo}. A FastAPI~\citep{fastapi} backend handles requests from the interface or external applications, builds the prompt from a system instruction, the retrieved context, and the user's question, and streams the generated response back to the interface. In the evaluation reported in Section~\ref{sec:evaluation}, the system instruction asked the model to answer directly and concisely, based strictly on the reference chapter, using standard aviation terminology, and to state explicitly when a question fell outside the scope of the chapter.

\begin{figure}[ht]
    \centering
    \includegraphics[width=0.74\textwidth]{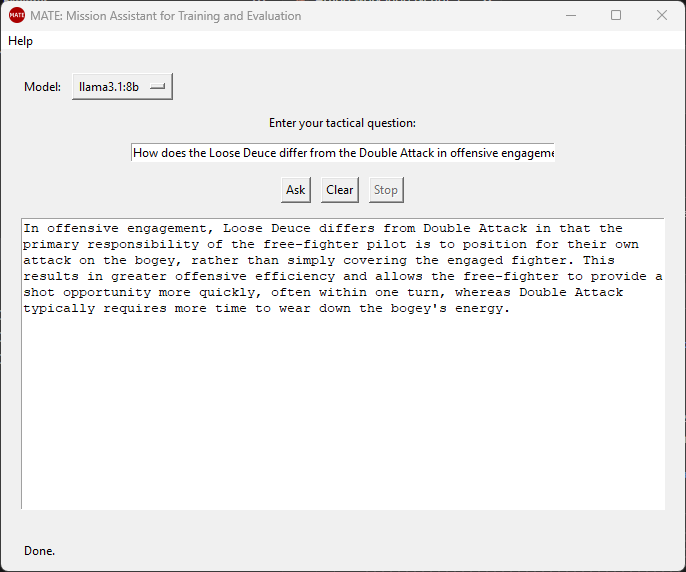}
    \caption{Screenshot of the MATE graphical interface.}
    \label{fig:ui_screenshot}
\end{figure}

\textbf{User Interface:} The interface is built using Tkinter~\citep{tkinter_docs}, a lightweight GUI library for Python. It allows users to type questions in natural language and receive AI-generated answers in real time, with a selector for the installed language models, Ask, Clear, and Stop controls, keyboard shortcuts, and error handling. All interactions with the backend occur through HTTP requests to the local FastAPI server. An optional voice input module based on Whisper~\citep{radford2023whisper} and PyAudio is also available, but it was disabled during the evaluation reported in Section~\ref{sec:evaluation}. Figure~\ref{fig:ui_screenshot} shows a screenshot of the user interface, where the assistant answers a tactical question using doctrinal content offline.

\begin{figure}[ht]
\centering
\begin{tikzpicture}[
    node distance=1.0cm and 0.5cm,
    base/.style={
        draw,
        rounded corners,
        minimum height=1.0cm,
        text width=2.1cm,
        align=center,
        font=\small
    },
    box/.style={base, fill=cyan!15},
    smallbox/.style={base, fill=gray!10},
    uibox/.style={base, fill=green!15},
    arrow/.style={
        thick,
        -{Latex[length=2mm]}
    }
]

\node[box] (docs) {Doctrinal\\Docs};
\node[box, right=of docs] (index) {Indexing\\(Embeddings\\+ FAISS)};
\node[box, right=of index] (retrieval) {Semantic\\Search};

\node[smallbox, above=of retrieval] (question) {Question};
\node[uibox, right=of question] (ui) {User\\Interface};

\node[smallbox, below=of retrieval] (cq) {Context +\\Question};
\node[smallbox, right=of cq] (llm) {LLM\\(Local API)};
\node[smallbox, right=of llm] (answer) {Answer};

\draw[arrow] (docs) -- (index);
\draw[arrow] (index) -- (retrieval);
\draw[arrow] (ui) -- (question);
\draw[arrow] (question) -- (retrieval);
\draw[arrow] (retrieval) -- (cq);
\draw[arrow] (cq) -- (llm);
\draw[arrow] (llm) -- (answer);
\draw[arrow] (answer) |- (ui);

\end{tikzpicture}
\caption{Overview of the offline tactical assistant system architecture.}
\label{fig:methodology_diagram}
\end{figure}
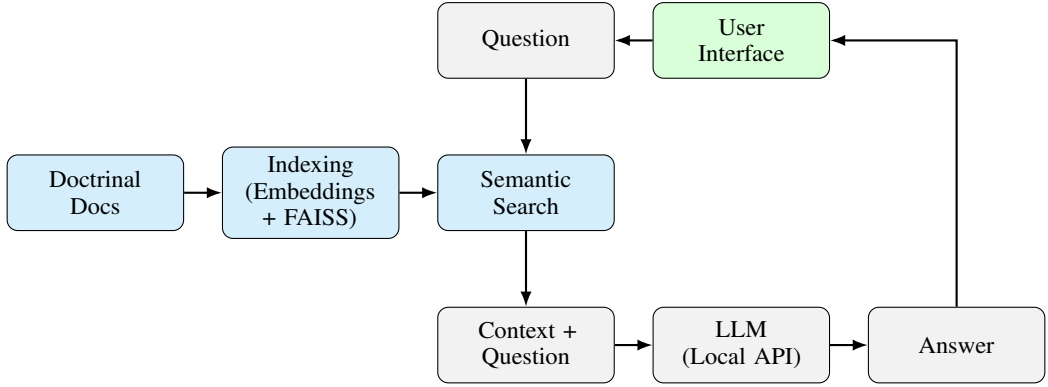

\textbf{Offline Deployment:} All components are designed to run entirely offline, supporting use in air-gapped environments. The embedding model, FAISS index, and the language model are all stored locally. The system can be packaged using tools like PyInstaller~\citep{pyinstaller} to create a standalone executable, and no internet connection is required at runtime. This ensures suitability for sensitive or classified military applications. Figure~\ref{fig:methodology_diagram} summarizes the overall architecture and data flow of the system, showing how user queries are processed through semantic retrieval and local language generation.

\section{Expert Validation}
\label{sec:evaluation}

To assess whether the proposed system can support tactical decision-making, an evaluation was conducted with Brazilian Air Force fighter pilots. As we cannot include or disclose classified military documents in this study, the knowledge base was built from a publicly available reference: Chapter 5, ``Section Tactics: Two-versus-One'', of Robert L. Shaw's manual ``Fighter Combat: Tactics and Maneuvering''~\citep{shaw1985fighter}. This chapter provides comprehensive descriptions of section-level tactics, two-versus-one engagements, and the tactical principles relevant to the evaluation.

Five questions were derived from the chapter to reflect typical queries a pilot or mission planner might have. They cover the benefits of mutual support and the evolution of section doctrine, from the Fighting Wing to the Double Attack and the Loose Deuce:

\begin{enumerate}[label=\textbf{Q\arabic*:}, leftmargin=*, itemsep=0em]
\small
    \item What is the primary advantage of flying as a section instead of a single fighter?
    \item What are the main advantages of the Fighting Wing formation for new pilots?
    \item What are the main weaknesses of the Fighting Wing doctrine?
    \item What tactical innovation did the Double Attack introduce compared to Fighting Wing?
    \item How does the Loose Deuce differ from the Double Attack in offensive engagement?
\end{enumerate}

Each question was submitted to the tool three times, once with each of the evaluated models, using the same indexed chapter and retrieval pipeline. The evaluation then followed three steps:

\begin{enumerate}[label=\textbf{\arabic*)}, leftmargin=*, itemsep=0em]
\small
    \item \textbf{Participant Profile}: Each pilot reported their operational experience, flight hours, and familiarity with LLMs.
    \item \textbf{Tactical Evaluation}: Each pilot rated the 15 answers (five questions, three models) on a 5-point Likert scale~\citep{boone2012analyzing}, from 1 (very inadequate) to 5 (very adequate).
    \item \textbf{General Evaluation}: Each pilot rated five tool-level aspects on the same scale and could leave free-text suggestions.
\end{enumerate}

The panel consisted of three fighter pilots with, on average, 6.7 years of operational experience and 933 flight hours. They reported moderate familiarity with LLMs (occasional use and basic knowledge), mostly through ChatGPT, Gemini, and Llama.

\begin{table}[ht]
\centering
\caption{Ratings of the answers to the tactical questions (mean $\pm$ standard deviation over three pilots, 1--5 scale).}
\label{tab:tactical}
\small
\begin{tabular}{lccc}
\toprule
Question & Llama 3.1 8B & Qwen 2.5 14B & Mistral NeMo 12B \\
\midrule
Q1 -- Advantage of flying as a section       & 4.67 $\pm$ 0.58 & 4.00 $\pm$ 1.00 & 4.67 $\pm$ 0.58 \\
Q2 -- Advantages of the Fighting Wing         & 4.67 $\pm$ 0.58 & 4.33 $\pm$ 0.58 & 4.67 $\pm$ 0.58 \\
Q3 -- Weaknesses of the Fighting Wing         & 3.00 $\pm$ 0.00 & 3.00 $\pm$ 0.00 & 3.00 $\pm$ 0.00 \\
Q4 -- Innovation of the Double Attack         & 3.33 $\pm$ 0.58 & 4.00 $\pm$ 1.00 & 4.67 $\pm$ 0.58 \\
Q5 -- Loose Deuce vs.\ Double Attack          & 5.00 $\pm$ 0.00 & 4.67 $\pm$ 0.58 & 5.00 $\pm$ 0.00 \\
\midrule
Average over questions                        & 4.13 & 4.00 & 4.40 \\
\bottomrule
\end{tabular}
\end{table}

\begin{table}[ht]
\centering
\caption{General evaluation of the tool (mean $\pm$ standard deviation over three pilots, 1--5 scale).}
\label{tab:general}
\small
\begin{tabular}{p{0.72\textwidth}c}
\toprule
Evaluation question & Rating \\
\midrule
\textbf{Relevance of retrieval}: did the information used in the answer correspond to the correct content from the reference material? & 3.67 $\pm$ 0.58 \\
\textbf{Conceptual and terminological accuracy}: did the model use correct tactical terms consistent with operational doctrine? & 4.33 $\pm$ 1.15 \\
\textbf{Integration of retrieved context}: did the model properly incorporate the retrieved text into its explanation? & 3.67 $\pm$ 0.58 \\
\textbf{Clarity and coherence}: was the answer clear, logical, and well structured? & 4.33 $\pm$ 0.58 \\
\textbf{Response time}: was the generation time appropriate compared to manually locating the information in the reference text? & 5.00 $\pm$ 0.00 \\
\bottomrule
\end{tabular}
\end{table}

Table~\ref{tab:tactical} shows the ratings of the answers to the tactical questions. All three models produced answers rated as adequate for most questions, with averages between 4.00 and 4.40. Mistral NeMo 12B obtained the highest average, matching or exceeding the other models in every question, while the largest difference between models appeared in Q4, where Llama 3.1 8B received 3.33 and Mistral NeMo 12B received 4.67. Q3, which asks about the weaknesses of the Fighting Wing doctrine, received a rating of 3 from every pilot for every model. Since this pattern does not depend on the model, it more likely reflects the content supplied by the retrieval step than the generation step itself.

This interpretation is consistent with the general evaluation in Table~\ref{tab:general}. The aspects related to retrieval, namely the relevance of the retrieved content and its integration into the answer, received the lowest ratings (3.67), whereas terminological accuracy and clarity were rated 4.33. Response time received the maximum rating from all pilots, indicating that querying the tool is faster than locating the information manually in the reference text. One factor that may affect retrieval is the context limit: with segments of 1,200 characters and a limit of 4,000 characters, only about three of the five retrieved segments reach the model in full, so relevant passages ranked lower may be truncated or left out.

These results should be interpreted as exploratory. The panel was small, the questions covered a single chapter of one manual, and the number of answers per model does not support statistical comparisons between models or a meaningful analysis of inter-rater agreement~\citep{koo2016guideline}. Nevertheless, they indicate that a fully offline RAG pipeline running on open-weight models can produce answers that experienced fighter pilots consider adequate for most tactical questions, and they point to retrieval as the main component to improve.

\section{Conclusion and Future Work}
\label{sec:conclusion}

This work presented MATE, an offline tactical assistant that combines document retrieval with a local language model to support decision-making and training in air combat operations. The tool allows users to ask questions in natural language and receive answers based on doctrinal manuals, even in environments without internet access. It was built using open-source tools and follows a modular architecture, making it suitable for integration into simulators and mission planning systems.

An evaluation with three Brazilian Air Force fighter pilots showed that the answers produced with three open-weight models were rated as adequate for most tactical questions, with Mistral NeMo 12B obtaining the highest average rating, and that the response time of the tool was considered fully adequate compared to manual search. The lowest ratings were associated with the relevance and integration of the retrieved content.

Future work will focus on five main improvements: \textbf{broader evaluation}, with more pilots, more questions, and additional tactical scenarios and manuals, enabling inter-rater agreement analysis; \textbf{retrieval quality}, improving how manuals are segmented, how relevant passages are selected, and how much context is passed to the model; \textbf{response time}, using more capable hardware to support larger models; \textbf{interactivity}, adding dialogue history to support follow-up questions and integrating the tool with simulation environments; and \textbf{voice interface}, optimizing the voice module, which is currently too heavy for routine use during flight or simulator-based training.

Offline tools like the one presented in this work may eventually help pilots access relevant tactical knowledge during training, operational missions, or post-mission debriefings. Although still in its early stages, this type of system has the potential to serve as a valuable support tool in simulators or onboard aircraft, particularly in air combat operations that demand quick decision-making under high pressure and without reliable connectivity.

\section*{Code Availability}

The source code of MATE is publicly available at \url{https://github.com/jpadantas/mate}. The repository includes the indexing, retrieval, backend, and interface components described in Section~\ref{sec:methodology}, together with the configuration used in the evaluation. The reference manual used in the evaluation is copyrighted and is not distributed with the code.

\bibliographystyle{abbrvnat}
\bibliography{ref}

\end{document}